# Artificial Intelligence Markup Language: A Brief Tutorial

Maria das Graças Bruno Marietto[1], Rafael Varago de Aguiar[1],
Gislene de Oliveira Barbosa[1], Wagner Tanaka Botelho[1], Edson Pimentel[1],
Robson dos Santos França[2], Vera Lúcia da Silva[3]

[1]Universidade Federal do ABC, São Paulo, Brazil
{graca.marietto,gislene.barbosa,wagner.tanaka,edson.pimentel}
@ufabc.edu.br
rafael.varago@aluno.ufabc.edu.br
[2]Tribunal Regional Eleitoral, São Paulo, Brazil
robson@robsonfranca.eti.br
[3]Instituto Federal de Educação, Ciência e Tecnologia de São Paulo, São Paulo, Brazil
verals@ifsp.edu.br

## ABSTRACT

*The purpose of this paper is to serve as a reference guide for the development of chatterbots implemented with the AIML language. In order to achieve this, the main concepts in Pattern Recognition area are described because the AIML uses such theoretical framework in their syntactic and semantic structures. After that, AIML language is described and each AIML command/tag is followed by an application example. Also, the usage of AIML embedded tags for the handling of sequence dialogue limitations between humans and machines is shown. Finally, computer systems that assist in the design of chatterbots with the AIML language are classified and described.*

## KEYWORDS

*Artificial Intelligence, Chatterbot, Pattern Recognition, Artificial Intelligence Markup Language (AIML), Tutorial.*

## 1. INTRODUCTION

In the current globalized society, the increasing development and spread of Information Technology and the Internet have led to the creation of distinct ways of communication among users in virtual environments. In this context, the cognitive interfaces provide new forms of interaction between human and machine. The Graphical User Interface (GUI) is based on navigation systems that use (i) hypertext and/or (ii) option selection with buttons/menus. These approaches require an additional cognitive effort by the users, since the natural language is the usual way of communication between humans. Thus, more appropriate cognitive interfaces allow the foundation of dialogues in natural language, and also make the human-machine interaction more attractive and powerful in terms of empathy.

In the context of Ubiquitous Computing, which goal is to effectively integrate human beings and technology/machine, research has been made regarding the development of the "natural interfaces". Such interfaces make the communication more intuitive, because the idea is to use usual forms of interaction such as natural language, gestures and vision. Among the possible forms of natural user interfaces, this paper highlights the usage of chatterbots or computer systems that are designed to simulate a conversation with humans [1]. In a chatterbot, conversation usually occurs by the exchange of text messages (as in a chat environment), and the chatterbot is organized to provide features that allow the usage of natural language, such as dealing of ambiguous situations and context-based message analysis (based on Pragmatics).

Currently, the chatterbots has been applied in the most varied purposes. For example, the entertainment chatterbots that are designed to entertain and to amuse the user while maintaining a coherent conversation. For instance, the chatterbot Susan is a virtual human that represents Dr. Susan Calvin from Isaac Asimov's Robot Serie [2]. There are chatterbots for mobile devices, such as the ChattyBot [3] for Android platform. Penny [4] is a virtual assistant to solve problems related to products and/or services offered by the Heat Gas Company Ltd. Also, educational chatterbots emerge as an alternative to the teaching-learning process and can be used to supplement the information provided in the classroom and to clarify any doubts. For instance, the Einstein chatterbot is responsible for to teach Physics and was developed by Artificial Life3 Company [5].

Along with the evolution of the chatterbots, researchers noticed that the structure of chatterbots modeled using the Natural Language Processing (NLP) would present itself as a complex task. In order to assist in such task, dedicated technologies for chattebot making has been developed. Among these technologies, Wallace [6] in collaboration with free software developers' communities can be noticed. From 1995 to 2000, the Artificial Intelligence Markup Language (AIML) is created, based on the concepts of Pattern Recognition, or Matching Pattern technique. It is applied to natural language modeling for the dialogue between humans and chatterbots that follow the stimulus-response approach. For this purpose a set of possible user inputs is modeled and, for each one of these sentences (stimuli), pre-programmed answers were built to be shown to the user. The ALICE (Artificial Linguistic Internet Computer Entity) chatterbot was the first to use the AIML language and interpreter. In ALICE, the AIML technology was responsible for pattern matching and to relate a user input with a response of the chatterbot's Knowledge Base (KB).

In the literature that presents the AIML concepts, there are tutorials that superficially introduce such concepts ([7]), or present the language concepts in detail ([6]). The two options are not suitable for AIML beginners because they cannot balance the amount of theory and application. Additionally, another works focus on the usage of Pattern Recognition techniques [8], and the AIML language itself is addressed as a secondary topic. In this work an AIML language tutorial is presented, concepts of Pattern Recognition are considered, but the main idea is to have a reference guide for AIML language's initial studies.

This paper is organized as follows. Section 2 presents a brief history of the development of chatterbots, followed by an introduction to Pattern Recognition. Section 3 introduces the AIML language and its main statements. Section 4 presents an application of AIML language for the treatment of conversations' limitations between people and machines. In Section 5 computational tools used for the development of chatterbots are classified and described. Lastly, Section 6 presents the conclusion s of the work.

## 2. CHATTERBOTS AND PATTERN RECOGNITION

The proposal to make dialogs with computers dates back the 50´s, when the British mathematician Alan Turing posed the question "Can machines think?" in the article "Computing Machinery and Intelligence" [9]. Turing proposed a test called the Imitation Game, now known as the Turing Test. The test is to make the interlocutor unable to distinguish whether he/she is talking to a person or a machine. In order to achieve it, the test is performed as follows:

- A digital computer, a human being and a judge (also human) are in different and isolated rooms;

- The judge establishes a dialogue with the computer and the human being, through a terminal, using a keyboard and video monitor;

- The computer passes the test if the judge is unable to tell whether the answers come from the human being or como from the computer.

Thus, the Turing Test can be considered the forerunner of today´s chatterbots. Laven [1] defines chatterbots as "... *a program that attempts to simulate typed conversation, with the aim of at least temporarily fooling the human into thinking they were talking to another person*". Chatterbots can be classified according to the techniques used in its development. Following this parameter, three generations can be identified. The first generation is characterized by the usage of NLP and Pattern Recognition´s basic techniques. In 1966 the ELIZA [10] chatterbot was developed by Joseph Weizenbaum and it poses as an early example of chatterbot design. During the 90's, the second generation chatterbots were built and the Artificial Intelligence (AI) techniques were applied, such as Artificial Neural Networks in conjunction with NLP techniques. JULIA chatterbot is an example of a second generation chatterbot developed by Michael Mauldin in 1994 [11]. The development of third-generation chatterbots uses more advanced Pattern Recognition techniques. The forerunner of this generation was the ALICE [6], developed in 2000 by Richard Wallace in partnership with the AliceBot community. The ALICE´s KB is implemented in AIML language. In [12] a current version of ALICE is available for testing.

This work chooses the chatterbots based on third generation techniques. The motivation is the fact that the usage of Pattern Recognition, in conjunction with AIML, presents features such as:

- Easy of implementation, since AIML is a XML (eXtensible Markup Language) based markup language, the tags make the implementation of dialogues easier (see sections 3 and 4);

- There are computational systems that help developers in the chatterbots´ codes creation and Web deployment for users access (see Section 5);

- There is a high level of reuse, since a significant amount of chatterbot projects is developed under a free/open software license. Therefore, the source code, documentation and examples are available for reuse, with suitable customizations for the needs of the new project. The ALICE is a prime example of a free software project, and other chatterbots were built based on ALICE. This reuse allows the design of systems without the need to build them from scratch.

## 2.1. Pattern Recognition for Chatterbots Modeling

Among the theories and technologies used to develop chatterbots, the Pattern Recognition is an example which aims to model computer systems based on the format of human dialogs. The Pattern Recognition is based on representative stimulus-response type blocks, which the user enters a sentence (stimuli) and the software makes an output (response) according to the user input. And in this sequence the dialogue carries on [6]. The language AIML is used to the development of chatterbots'KB, if the chatterbot adopts the Pattern Recognition technique.

The ALICE was the first chatterbot with an AIML language implemented KB. Its operation is divided in three steps. The first step is the question input by the user. In the second step the system (i) performs word processing actions to fit the user's input to a pre-established format by the designer and (ii) makes the pattern matching between user input and the KB. Finally, one answer is presented to the user in the third step [6].

# 3. AIML LANGUAGE: SYNTAX AND SEMANTIC

The AIML language´s purpose is to make the task of dialog modeling easy, according to the stimulus-response approach. Moreover, it is a XML-based markup language and it is a tag-based. Tags are identifiers that are responsible to make code snippets and insert commands in the chatterbot. AIML defines a data object class called AIML objects, which is responsible for modeling patterns of conversation [6]. Technically speaking, AIML objects are language tags, and each tag corresponds to a language command. The general form of an AIML object/command/ tag has the following structure:

```
<command> ListOfParameters </command>
```

An AIML command consists of a start tag (`<command >`), a closing tag (`</command >`) and a text (`ListOfParameters`) that contain the command's parameter list. AIML is an interpreted language. As such, each statement is read, interpreted and executed by a software known as interpreter. Section 5 presents some AIML interpreters that are available on the Internet.

AIML is based on basic units of dialogue, formed by user input patterns and chatterbot responses. These basic units are called categories, and the set of all categories makes the chatterbot KB. Among the AIML objects, the following tags are worth citing: `category`, `pattern` and `template`. The `category` tag defines a unit of knowledge (of dialogue) of the KB. The `pattern` tag defines a possible user input, and the `template` tag sets the chatterbot response for a certain user input.

This section is structured as follows: Subsection 3.1 presents the AIML vocabulary and Subsection 3.2 details a subset of AIML tags.

## 3.1. Language´s Vocabulary

The AIML vocabulary consists of words, spaces and the special characters "*" and "_", known as wildcards. Wildcards are used to replace a string (words or sentences). The AIML interpreter gives higher priority to categories containing patterns that use the wildcard "_" than "*". Then, the categories with the wildcard "_" are analyzed first. For an AIML object/tag be well defined, it must follow the XML standards. For example, object names cannot start with numbers, they are case-sensitive (there is a distinction between uppercase and lowercase letters) and blanks are not allowed.

In this work, the text into `<pattern>` tag is written in capital letters in KB. This standard helps in the stage of standardization and simplification, responsible for modifying the user's text in order to standardize the sentences. For example, typing "Please inform how the pattern command works", the sentence is replaced by the question "PLEASE INFORM HOW THE PATTERN COMMAND WORKS". The transformation for capital letters simplifies the search task, since all characters are in the same format.

## 3.2. Language´s *Tags*

In this section a subset of AIML tags is presented.

### `<aiml>` Tag

Table 1 shows the AIML code that illustrates a usage of the `<aiml>` tag. Each AIML file begins with `<aiml>` tag and is closed by a `</aiml>` tag, in Line 1 and Line 8, respectively. This tag contains the `version` and `encoding` attributes, as defined in Line 1. The `version` attribute identifies the AIML version used in the KB. This example uses the version 1.0.1. If this attribute is omitted, it will not result in program errors, but may cause confusion during maintenance tasks or system upgrade. The `encoding` attribute identifies the type of character encoding that will be used in the document. In this example, UTF-8 is used. Within the scope `<aiml> </ aiml>` (lines 1-8) at least one element `<category>` must exist.

Table 1.  Example of AIML Code.

```
1  <aiml version="1.0.1" encoding="UTF-8"?>
2  <category>
3     <pattern> HELLO BOT </pattern>
4     <template>
5        Hello my new friend!
6     </template>
7  </category>
8  </aiml>
```

## `<category>` Tag

The basic units of an AIML dialog are called categories. Each category is a fundamental unit of knowledge contained in the chatterbot KB. A category consists of (i) an user input, in the form of a sentence (assertion, question, exclamation, etc), (ii) a response to user input, presented by the chatterbot, and (iii) an optional context.

A KB written in AIML is formed by a set of categories. The categories are organized by subjects and stored in files with `.aiml` extension, in order to tidy things up and ease the knowledge base´s maintenance process. Category modeling is made by using the `<category>` and `</category>` tags. Table 1 shows a sample code for the `<category>` tag that correspond to the following dialogue excerpt between the user and the chatterbot:

```
User: Hello bot
Bot: Hello my new friend!
```

In Table 1, lines 2 and 7 present the opening and closing of the `<category>` tag, respectively. The categories must be inside the `<aiml> </aiml>` context (lines 1-8) and must contain a `<pattern>` tag (Line 3) and a `<template>` tag (Line 4). The information entered by the users is bounded in the `<pattern>` tag. Also, the answers provided by the chatterbot are in `<template>` tag. Thus, in Line 3 the category understands that a possible user input can be "HELLO BOT". For this stimulus, the chatterbot will answer the user with the sentence "Hello my new friend!" (see lines 4-6).

## `<pattern>` Tag

The `<pattern>` tag contains a possible user input. There is just a single `<pattern>` in each `<category>` tag, and it must be the first element to be set. Also, the words are separated by single spaces, and wildcards can replace parts of a sentence. In Table 1 the AIML code `<pattern>` HELLO BOT `</pattern>`, defined in Line 3, indicates that the chatterbot developer understood that the sentence "HELLO BOT" is a possible user input.

## `<template>` Tag

The `<template>` tag contains possible chatterbot answers to the user. It must be within the scope of a `<category>` tag, and be placed after the `<pattern>` tag. Most of the chatterbot information is bounded by this element. This tag can save data and activate other programs, or even give conditional answers and call other answers´ categories.

In AIML code presented in Table 1, the `<template>` tag provides a response to the user, which in this case only displays the sentence "Hello my new friend!", as shown in lines 4 to 6.

## `<star index = "n"/>` Tag

The `<star index=""n"/>` tag captures a particular text fragment, contained in the user input sentence. The index `n` indicates the phrase component that will be mapped and captured. Thus, it is observed that:

- `<star index="1"/>`: equivalent to first fragment of the text;

- `<star index="2"/>`: equivalent to second fragment of the text;

- `<star index="3"/>`: equivalent to third fragment of the text;

- And so on.

With this command is possible to store user text snippets. The attribute `"n"` in `index = "n"` is optional and if it is omitted the value 1 is assumed. Thus, the tag `<star/>` is the same as `<star index="1"/>`.

The AIML code of Table 2, lines from 1 to 6, models the following generic dialog sequence:

```
User: I LIKE *
Bot: I like * too.
```

From this template it is possible to the chatterbot answer that likes anything that the user reports that he/she likes. Here is a possible conversation between user and bot:

```
User: I like video games
Bot: I like video games too.
```

This conversation is possible using the `<star/>` tag and the wildcard "`*`". In Line 2 of Table 2 the information about the user preferences is identified by the "`*`" wildcard. In Line 4 the `<star/>` command (without index `n`) captures the contents identified by wildcard in Line 2, and put this content in the answer to be sent to the user.

Table 2. `<star>` Tag Usage Example.

```
1    <category>
2       <pattern> I LIKE * </pattern>
3       <template>
4          I like <star/> too.
5       </template>
6    </category>
7
8    <category>
9       <pattern> A * IS A * </pattern>
10      <template>
11         When a <star index="1"/> is not a <star index="2"/>?
12      </template>
13   </category>
```

Table 2 also exemplifies the usage of `<star>` tag in the code from line 8 to 13, corresponding to the following generic dialogue snippet:

```
User: A * IS A *
Bot: When a * is not a *?
```

From this generic interaction model, the following dialogue is possible between user and bot:

```
User: A rose is a flower
Bot: When a rose is not a flower?
```

In Line 9, when a possible user input is defined, two wildcards are used. In the example of the dialogue about the rose, the first wildcard identifies the word "ROSE" and the second identifies the word "FLOWER". In Line 11 the instructions `<star index="1">` and `<star index="2">` present the words "ROSE" and "FLOWER", respectively.

## `<srai>` Tag

One of the most useful properties of AIML language is the possibility to target different `<pattern>` models (user input) for a single `<template>` tag (chatterbot response). Thus, AIML interpreter can efficiently look for an answer from different types of user input. This possibility is obtained by using the `<srai>` tag.

The following is presented a description of some `<srai>` tag applications as described in [6]: symbolic reduction, divide and conquer, synonyms resolution and keywords detection.

### Symbolic Reduction with the `<srai>` Tag

The symbolic reduction technique is used for pattern simplification. Thus, a complex grammar pattern is mapped into a set of simpler patterns. For instance, the question "WHO IS X", where X represents any entity, could be written in several ways, such as: "DO YOU KNOW WHO X IS" and "TALK MORE ABOUT X". The `<srai>` tag is used to map among these standards. Table 3 presents the AIML code with an example of the usage of the symbolic reduction technique.

Table 3. Symbolic Reduction with `<srai>` Tag.

```
1   <category>
2      <pattern> WHO IS ALAN TURING? </pattern>
3      <template>
           Alan Turing was a British mathematician, cryptographer,
4          and computer scientist often credited as
           the founder of modern Computer Science.
5      </template>
6   </category>
7
8   <category>
9      <pattern> WHO IS ALBERT SABIN? </pattern>
10     <template>
           Albert Sabin was the researcher who developed
11         the vaccine that is the main defense against polio.
12     </template>
13  </category>
14
15  <category>
16     <pattern> DO YOU KNOW WHO * IS? </pattern>
17     <template>
18         <srai> WHO IS <star/> </srai>
19     </template>
20  </category>
```

In the code of Table 3 there is a category prepared to talk about Alan Turing (lines 1-6), and another category prepared to talk about Albert Sabin (lines 8-13). The flowchart in Figure 1 illustrates the `<pattern>` commands of these categories in A and B, respectively. The first category identifies the user's question "WHO IS ALAN TURING?", and the second category identifies the question "WHO IS ALBERT SABIN?".

Considering that the user can ask about the same researchers in different ways, the application of the symbolic reduction technique begins with the creation of a new category that models a varied way of asking about the researchers. This category is encoded in lines 15 to 20 and it considers that the user may ask the question as follows: "DO YOU KNOW WHO * IS?", where the wildcard "*" identifies the name of the person that the user wants to search. If the user enters a text that matches the pattern defined in Line 16 (corresponding to C in Figure 1), the command `<srai> WHO IS <star/> </srai>` (Line 18) redirects the chatterbot response for another category. The `<star/>` command inserts the text captured by the wildcard in Line 16. In D a test is made in order to check what value is stored in `<star/>`: ALAN TURING or ALBERT SABIN. In E and F the category responsible for talking about these people is called.

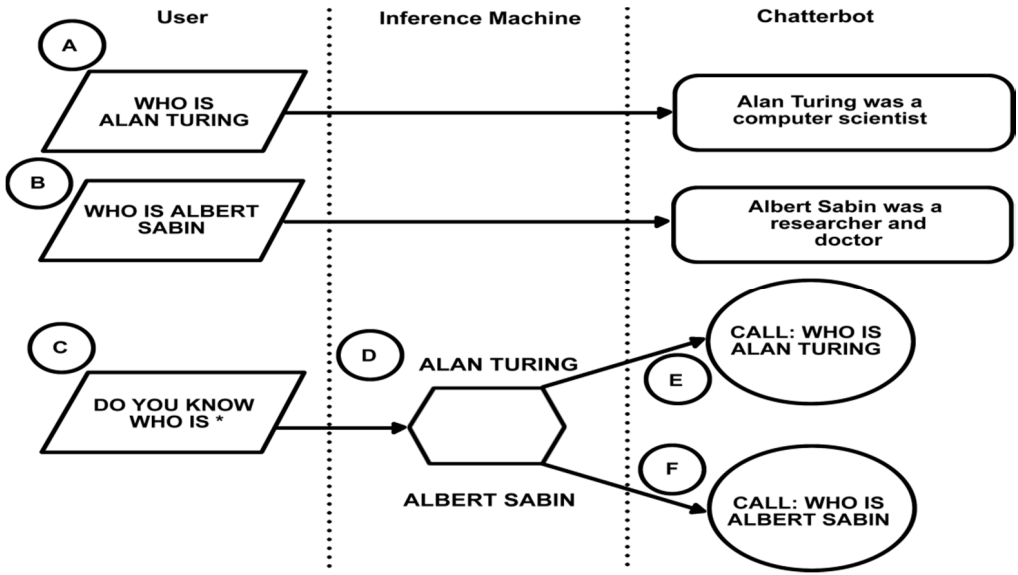

Figure 1. Flowchart of Symbolic Reduction Technique.

## Divide and Conquer with the `<srai>` Tag

According to [6], "*many individual sentences may be reduced to two or more sub sentences, and the reply formed by combining the replies to each*". For example, the sentence "Bye X", begins with the word "Bye" followed by X, where X is any sequence of characters. The proposal of divide-and-conquer technique is that a user input that begins with the word "Bye", and followed by any string is treated by the `<pattern>BYE</pattern>` command.

Table 4. Divide and Conquer with `<srai>` Tag.

```
1   <category>
2      <pattern> BYE </patter>
3         <template> Goodbye friend! </template>
4   </category>
5
6   <category>
7      <pattern> BYE * </pattern>
8      <template> <srai> BYE </srai> </template>
9   </category>
```

Figure 2 and Table 4 present the AIML code of a conversation that uses this technique. Line 7 of Table 4 and item A of Figure 2, define a user input pattern with the following feature: a form of saying goodbye starting with the "Bye" word, followed by the "*" wildcard representing any sentence. In this case, in Line 8 of Table 4 (Item B of the flowchart in Figure 2), chatterbot redirects the execution flow system for Line 2 with the `<srai> BYE </ srai>` command.

Line 2 and C define the pattern "BYE". Thus, the response in Line 3 and D will be presented to user.

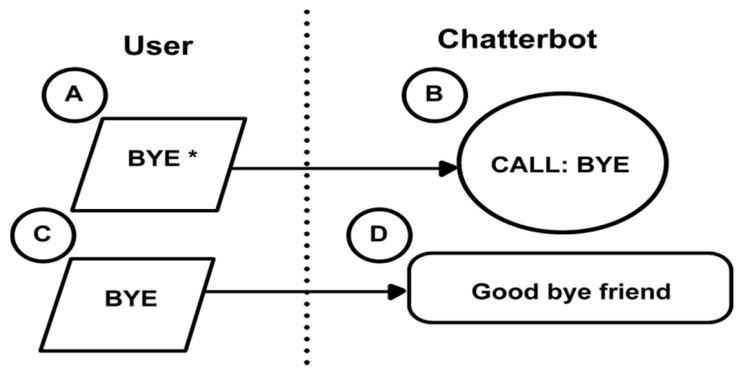

Figure 2. Flowchart of Divide-and-Conquer Technique.

## Synonyms Resolution with the `<srai>` Tag

The treatment of synonyms is important because in a conversation it is possible to appear different words with similar meanings, depending on the context. Using `<srai>` tag the chatterbot presents a same answer for synonyms, by mapping a set of inputs related to a root pattern.

Table 5 illustrates the use of the synonyms resolution technique. In this example, the words "INDUSTRY" (Line 2) and "FACTORY" (Line 9) are treated as synonyms. If the user enters the word "INDUSTRY", the chatterbot will present the response defined in lines 3-5. The following conversation snippet will be established between user and bot:

```
User: Industry
Bot: It is a development center.
```

If the user enters the word "FACTORY", the chatterbot will redirect the response by using the `srai> INDUSTRY </srai>` command in Line 11, to the pattern defined in Line 2. Consequently, the answer defined in rows 3 to 5 will be presented to the user, resulting in the following iteration:

```
User: Factory
Bot: It is a development center.
```

Table 5. Synonyms Resolution with `<srai>` Tag.

```
1   category>
2      <pattern> INDUSTRY </pattern>
3      <template>
4         It is a development center.
5      </template>
6   </category>
7
8   <category>
9      <pattern> FACTORY </pattern>
10     <template>
11        <srai> INDUSTRY </srai>
12     </template>
13  </category>
```

## Keyword Detection with `<srai>` Tag

It is possible to define the same response when a specific keyword is found in the user input. Such response is obtained regardless of the relative position of the word in the sentence. For example, if "FAMILY" is a keyword, it is possible to define a set of categories responsible to identify its occurrence in a particular sentence and, after this identification, to direct the chatterbot´s answer to a unique `template` command.

The AIML code of Table 6 illustrates an application of the keyword detection technique. The patterns presented in lines 2, 9, 16 and 23 identify all sentences that contain the word "FAMILY" in its structure: FAMILY (Line 2), _FAMILY (Line 9), FAMILY* (Line 16) and _FAMILY (Line 23). Observe the combined use of special wildcard characters "*" and "_".

Table 6.  Keyword Detection with `<srai>` Tag.

```
1   <category>
2      <pattern> FAMILY </pattern>
3      <template>
4         Family is an important institution.
5      </template>
6   </category>
7
8   <category>
9      <pattern> _ FAMILY </pattern>
10     <template>
11        <srai> FAMILY </srai>
12     </template>
13  </category>
14
15  <category>
16     <pattern> FAMILY * </pattern>
17     <template>
18        <srai> FAMILY </srai>
19     </template>
20  </category>
21
22  <category>
23      <pattern> _ FAMILY * </pattern>
24      <template>
25         <srai> FAMILY </srai>
26      </template>
27  </category>
```

The keyword detection technique defines that, for any input sentence with the keyword "FAMILY", the redirection execution flow will occur to the pattern defined in Line 2. It is performed by `<srai> FAMILY </srai>` in lines 11, 18 and 25. Thus, the response to be sent to the user will be "Family is an important institution", defined in lines 3 to 5.

Considering the code in Table 6, the conversation excerpts between the user and the chatterbot are presented:

```
User: I love my family
Bot: Family is an important institution.

User: I don´t have a family
Bot: Family is an important institution.
```

## `<random>` and `<li>` Tags

The `<random>` tag is used to respond to a user input in different ways. Each possible response must be delimited by `<li>` tag. In this way, the chatterbot responses are handled as a list, and the answers are randomly selected by the AIML interpreter.

Table 7 presents the AIML code with an example of random answers, where to the user input "HI" (see Line 2) the answer is randomly selected (`<random>` tag, Line 4) among the three responses mapped by `<li>` tags in lines 5 to 7.

Table 7. `<random>` and `<li>` Tags Usage Example.

```
1   <category>
2      <pattern> HI </pattern>
3      <template>
4         <random>
5            <li> Hi! Nice to meet you </li>
6            <li> Hello, How are you? </li>
7            <li> Hello! </li>
8         </random>
9      </template>
10  </category>
```

The following conversation excerpts may occur, considering the AIML code in Table 7:

```
User: Hi
Bot: Hello!

User: Hi
Bot: How are you?
```

## `<set>` and `<get>` Tags

The `<set>` and `<get>` AIML tags allow the chatterbot to work with variables. There are predefined variables in the language, to store information such as data about the chatterbot (*e.g.*, name, gender, location). However the variables can also be created by programmers, by defining a name and initialization values.

The `<set>` tag is used to create variables, and must be within the scope of the `<template>` tag. Its syntax is as follows:

<center>`<set name = "variableName"> variableValue </set>`</center>

Where `variableName` is the name of the variable to be created or changed, and the `variableValue` is the value to be assigned to the variable.

Table 8 illustrates an example of using the `<set>` tag. In this case, the name informed by user in the sentence "MY NAME IS *", Line 2, is identified by "*" wildcard. After this, in Line 4 the `<set name= "nameUser"> <star/> </set>` command stores, in the variable `nameUser`, the text captured by the `<star/>` command, which in this example is the name of the user.

Table 8. `<set>` Tag Usage Example.

```
1   <category>
2      <pattern> MY NAME IS * </pattern>
3      <template>
4         Hello <set name="nameUser"> <star/> </set>
5      </template>
6   </category>
```

The `<get>` tag returns the value stored by `<set>` tag, and should also be within the scope of the `<template>` tag. Its syntax follows the form:

<center>`<get name = "variableName"/>`</center>

The code illustrated in Table 9 presents the use of `<get>` tag. Pursuing the logic of the code of Table 8, now in Table 9 the value previously stored in `nameUser` is used to respond to the user input "GOOD NIGHT", defined in `<pattern>` tag in Line 2. This sentence is answered with the contents of `<template>` Good night `<get name="nameUser"/>` `</template>` command. Note that `<get name="nameUser"/>` command presents to the user the value stored in variable `nameUser`.

Table 9. `<get>` Tag Usage Example.

```
1   <category>
2       <pattern> GOOD NIGHT </pattern>
3       <template>
4           Good night <get name="nameUser"/>
5       </template>
6   </category>
```

Considering AIML code in tables 8 and 9, the following conversation may occur between the user and bot:

```
User: My name is João
Bot: Hello João
User: Good night
Bot: Good night João
```

## `<that>` Tag

The Pragmatic is a branch of Linguistic that studies the language in social contexts, and how the contexts contribute to comprehend meanings. Through the pragmatic analysis of sentences it is possible to infer the direction of the conversation, taking into account the sentence and the context where the dialogue is established. To deal with the Pragmatics issue in chatterbots development, AIML offers the `<that>` and `<topic>` tags.

The `<that>` tag tells the system to analyze the last sentence presented by the chatterbot. It is noteworthy that to analyze the latest sequence of chatterbot is important when the bot has held a question, and the user's response needs to be contextualized in relation to this question. This tag must be within the `<category>` scope.

Table 10 shows an example of AIML code that makes use of `<that>` tag. In this example, the possible answers, "YES" or "NO" (defined in the patterns of lines 9 and 17, respectively), can only be properly understood when compared with the last question asked by the chatterbot: "Do you like movies?". It is asked in lines 3 to 5. The relationship between the chatterbot´s question "Do you like movies?" and the user responses "YES" or "NO" is done in lines 10 and 18, with the use of `<that>` tag.

It is worth noting that when a category uses `<that>` tag, the chatterbot's answer will be displayed only if the AIML interpreter validate the `<that>` content.

Following a possible conversation between the user and the bot is presented, considering the code in Table 10:

```
User: Make some question
Bot: Do you like movies?
User: No
Bot: OK. But I like movies.
```

Table 10. `<that>` Tag Usage Example.

```
1   <category>
2       <pattern> MAKE SOME QUESTION </pattern>
3       <template>
4           Do you like movies?
5       </template>
6   </category>
7
8   <category>
9       <pattern> YES </pattern>
10      <that> Do you like movies? </that>
11      <template>
12          Nice, I like movies too.
13      </template>
14  </category>
15
16  <category>
17      <pattern> NO </pattern>
18      <that> Do you like movies? </that>
19      <template>
20          OK. But I like movies.
21      </template>
22  </category>
```

## `<topic>` Tag

The `<topic>` tag is used to organize subjects/topics that the chatterbot will be able to talk. To achieve this, the categories that deal with the same subject are grouped together, to improve the search for reasonable chatterbot's responses, to be sent to user. This tag allows the simulation of an important feature of dialogues between humans, namely: to lead the conversation to a specific subject, talk about this subject, and then identify when there is a change of subject during the conversation.

Thus, chatterbot can have a generic subject/topic, defined as `<topic> * </ topic>`, or a root subject/topic like `<topic> HELLO </ topic>`. In addition to these generic subjects it is possible, with the `<topic>` tag, to group categories related to themes such as food, philosophy, yoga, etc. In turn, the seek time of the AIML interpreter decreases due to the fact that the focus for patterns searching is restricted to categories contained in `<topic>` scope.

Table 11 illustrates an AIML code to exemplify the usage of `<topic>` tag. In this example, the chatterbot is modeled to talk about flowers, and it is defined in the `<topic name="flowers"> </topic>` scope in lines 8 to 22. The direction flow of the chatterbot for categories contained in the `flowers` topic occurs whenever the `topic` variable gets `flowers` value through the `<set name="topic">flowers</set>` command, as shown in Line 4. After that, the next pattern detected by the chatterbot will be initially perceived in categories of `flowers` topic. In this example, there are two categories in `flowers` topic. The first is to work with a generic user input (see Line 10), and the second is to work with the sentence "I LIKE IT SO MUCH!", as defined in Line 17.

The following excerpt of a possible conversation between the user and the bot is presented, considering the code in Table 11:

```
User: Let talk about flowers.
Bot: Yes.
User: Rose is my favourite flower
Bot: Flowers have a nice smell.
User: I like it so much!
Bot: I like flowers too.
```

Table 11. `<topic>` Tag Usage Example.

```
1  <category>
2     <pattern> LET TALK ABOUT FLOWERS. </pattern>
3     <template>
4        Yes <set name="topic">flowers</set>
5     </template>
6  </category>
7
8  <topic name="flowers">
9     <category>
10       <pattern> * </pattern>
11       <template>
12          Flowers have a nice smell.
13       </template>
14    </category>
15
16    <category>
17       <pattern> I LIKE IT SO MUCH! </pattern>
18       <template>
19          I like flowers too.
20       </template>
21    </category>
22 </topic>
```

## **`<think>` Tag**

The content of `<think>` tag is processed by the chatterbot, but not displayed to user. This tag is used for data processing, conditional statements and tests that should not be visible to user. In Table 12, an AIML code example of this tag is displayed. In Line 4 the chatterbot stores the user name in `nameUser` variable, without the user having knowledge of this assignment, once the `<set>` command is in the `<think>` tag scope.

Table 12. `<think>` Tag Usage Example.

```
1  <category>
2     <pattern> MY NAME IS * </pattern>
3     <template>
4        <think> <set name="nameUser"> * </set> </think>
5     </template>
6  </category>
```

## **`<condition>` Tag**

The `<condition>` tag is used whenever (i) there is a list of possible answers to be presented to the user, and (ii) the choice of the most appropriate response relies on the analysis of a particular variable that was updated during the conversation between the user and the chatterbot. The `<condition>` semantics is equivalent to the `case` command semantics, found in many programming languages. This tag takes as parameters the variable name and the value to be compared, as follows:

```
<condition name="variableName" value="variableValue"/>
```

Where `variableName` is the name of the variable to be checked, and `variableValue` represents the value that must be compared. If the value matches, the code block delimited by `<condition>` tag is executed.

Table 13 presents an example of the usage of the `<condition>` tag. The `state` variable is used to store the user's state of mind and may take the values `happy` and `sad`. The assignment of these values is made during the conversation between user and chatterbot. Line 4 verify if state is equal to `happy`. If equal, the chatterbot shows the sentence "It is nice being happy" to the user, as defined in Line 5. In Line 7 is checked if `state` is equal to `sad`. If equal, chatterbot shows to the user the sentence "Being sad is not nice", as defined in Line 8.

Table 13. `<condition>` Tag Usage Example.

```
1   <category>
2      <pattern> HOW ARE YOU? </pattern>
3      <template>
4         <condition name="state" value="happy">
5            It is nice being happy.
6         </condition>
7         <condition name="state" value="sad">
8            Being sad is not nice.
9         </condition>
10     </template>
11  </category>
```

## `<bot>` Tag

AIML allows developer to define some chatterbot´s properties with `<bot>` tag, and such properties can be seen by the user during the conversation. Table 14 presents an AIML code using this tag. In this example, when the chatterbot detects the user input "BOT'S PROPERTIES", as defined in Line 2, it shows some of its features (see lines 3-11).

Table 14. `<bot>` Tag Usage Example.

```
1   <category>
2      <pattern> BOT'S PROPERTIES </pattern>
3      <template>
4         <bot name="age"/>
5         <bot name="gender"/>
6         <bot name="location"/>
7         <bot name="nationality"/>
8         <bot name="birthday"/>
9         <bot name="sign"/>
10        <bot name="botmaster"/>
11     </template>
12  </category>
```

# 4. USE OF AIML TAGS IN TREATMENT OF LIMITATIONS IN SEQUENCE OF CONVERSATION WITH CHATTERBOTS

This section presents the integrated use of AIML tags for the treatment of ambiguity. The goal is to illustrate how AIML tags, shown in the Section 3, can be used to model and implement efficient and robust chatterbots. The computer systems that work with natural languages should be able to deal with issues such as ambiguity, treatment of synonyms, conversation's intention analysis that depends on the context (Pragmatics), and so on.

Linguistic defines ambiguity as the characteristic of some terms or expressions to have a double meaning, that is, more than a possible understanding [13]. Thus, chatterbots need to identify ambiguous words and direct the conversation flow to the correct category. In order to reduce the aggravating factors of ambiguity in conversations between human and chatterbots, this paper proposes a mechanism called "One Step Reverse Resolution". In such mechanism, a conditional test is performed in a control variable, and the test result is used to lead the conversation flow the right category. A control variable is defined for each ambiguous category, and its value identifies the dialog's intention. Therefore, in an ambiguous user input, the value of the control variable is checked and the real user intention with some degree of truth can be inferred based on such value.

Table 15 shows two categories with the word "LIE" in the sentence of the `<pattern>` tag. The category implemented in lines 1 to 7 treats "LIE" as "mislead". On the other hand, the category defined in lines 9 to 15 treats "LIE" as "down". To address this ambiguity issue the `control_lie` variable is defined, and may take the values `not_true` or `not_stand`, depending on the conversation contex. These values are assigned during the conversation with the user. When the chatterbot detects an ambiguous keyword, it uses the `control_lie` value to establish the user intention.

Table 15.  Example of Ambiguity Treatment.

```
1   <category>
2       <pattern> I TOLD YOU THE TRUTH </pattern>
3       <template>
4           Thank you!
5           <think> <set name="control_lie"> not_true </set> </think>
6       </template>
7   </category>
8
9   <category>
10      <pattern> I WANT TO SLEEP </pattern>
11      <template>
12          Good night!
13          <think> <set name="control_lie"> not_stand </set> </think>
14      </template>
15  </category>
16
17  <category>
18      <pattern> WHAT DO YOU THINK ABOUT LIE? <pattern>
19      <template>
20          <condition name="control_lie" value="not_true">
21              Why did you lie to me?
22          </condition>
23          <condition name="control_lie" value="not_stand">
24              It is better you go to bed.
25          </condition>
26      </template>
27  </category>
```

In Figure 3 the flowchart illustrates this scenario. A, B and C in flowchart (lines 1-7 in Table 15) indicate that, if the user enters the sentence "I TOLD YOU THE TRUTH", the chatterbot will assign the value `not_true` to `control_lie`, and will present the sentence "Thank you" to the user. On the other hand, D, E and F (lines 9-15) indicate that, if the user enters the sentence "I WANT TO SLEEP", the chatterbot will assign the value `not_stand` to `control_lie`, and will submit the sentence "Good night" to the user.

In Table 15, the ambiguity treatment is implemented in the lines 17 to 27. Also, it is depicted in G, H and I in Figure 3. In this example, if the user enters the sentence "WHAT DO YOU THINK ABOUT LIE?" (as defined in Line 18), the answer presented to the user will depend of the `control_lie` variable current value. If `control_lie` is `not_true` (Line 20 in Table 15, Item H in Figure 3), the bot will show the sentence "Why did you lie to me?" to the user. If it is `not_stand` (Line 23 in Table 15, Item H in Figure 3), the bot will show the phrase "It is better you go to bed." to the user.

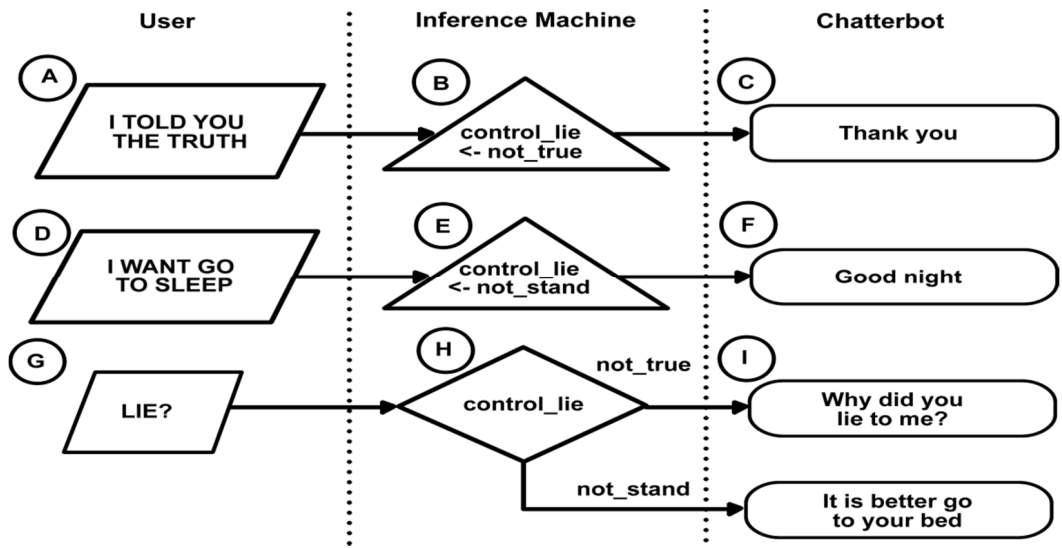

Figure 3. Flowchart of Ambiguity Treatment.

## 5. COMPUTER SYSTEMS FOR CHATTERBOTS IMPLEMENTATION

The AIML is an interpreted language. Interpretation is one of the processes by which the source code of a program is translated into the machine code, to be able to run on a computer. In the interpreting process, the interpreter program repeatedly performs the following sequence, for each source code line: (1) it captures the program's source code line; (2) it checks and analysis the line, both syntactically and semantically; (3) if there are no errors, it translates to a binary language for further execution.

The computer systems that assist the chatterbot implementation, based on the AIML language, have an interpreter in its structure. However, these systems differ and can be classified according to the following categories: AIML editors and chatterbots' development platforms. In the next subsections, some examples of such software are introduced and described.

### 5.1. AIML Editors

AIML editors serve as test bed environments for AIML commands and the logic of conversations sequence. Using AIML editors the code can be debugged and enhanced while they are built by chatterbot's developers. These test bed environments decrease the development time, and save efforts with software installation and configuration. Among the editors available in the literature it is worth citing:

- GaitoBot [14]: this is an offline tool. In other words, it needs to be installed in a computer to be used. It is implemented in C# language and it is best advised for testing of AIML code snippets. Also, it provides a set of graphical components (buttons, labels, etc.) that helps in the code writing process. Rather than writing every AIML command, GaitoBot provides to the chatterbot developer a set of graphical components that represent AIML tags. At the end of this process, the GaitoBot interpreter generates an `.aiml` file that contain the AIML code that corresponds to the graphical components that are used;

- Simple AIML Editor [15]: this tool provides an offline AIML interpreter, enclosed in an attractive user interface. It is implemented in C# language, and the Simple AIML Editor (i) allows the user to check the validity of AIML code and (ii) provides the comment manager, a mechanism to help the botmaster to write comments at the beginning of each AIML file. The botmaster is the individual who manages the chatterbot to perform actions such as analysis of the log conversations, the detection of errors and inconsistencies in the KB, the creation of new rules on questions that are not answered by bot, among others.

## 5.2. Development Platforms

A chatterbot development platform provides the AIML interpreter, as well as a set of integrated technologies, to reduce the cost and the complexity of development, deployment and management of such systems. Some of technologies that can be pointed out are the database integration, Web services for an online chatterbot and the log recording of the conversations with the users. Some development platforms are presented below:

- ProgramD [16] consists of an API (Application Programming Interface) implemented in Java. Among the advantages of using ProgramD there are (i) the distribution of the source code files and (ii) a KB that can be used for testing and analysis during the AIML language learning, as well as to develop chatterbots;
- Program# [17] is a .NET platform AIML that allows the usage of programming languages targeted to the .NET platform, such as C#, VB.NET, F#, and so on. Program# enables the chatterbot integration with Microsoft Windows and Web systems. Since the tool is distributed as a DLL (Dynamic Link Library) it is possible to attach it in encapsulated form project.

## 6. CONCLUSIONS

This work highlighted the importance of the conversational software development, more specifically the chatterbots as computational agents to establish natural language dialogs between humans and machines. Among the chatterbot development theories, the Pattern Recognition was outlined, and it is based on representative stimulus-response blocks. The AIML language is one of the most widely used technologies for chatterbots implementation that combines the technical and theoretical Pattern Recognition's infrastructure in its development.

In order to provide to the developers a reference guide to direct them in the chatterbots development, this work introduced the main tags (commands) of AIML so the reader could be able to start developing chatterbots, to build his/her own KB and to integrate it with the programming language of his/her choice, such as Java or C#.

## REFERENCES

[1] S. Laven (2013), "The Simon Lave Page", Available at: http://www.simonlaven.com, Accessed on 10th January 2013.

[2] SusanCalvin.org (2013), "SusanCalvin: The Robopsychologist", Available at: http://www.susancalvin.org/index.php/talk-with-susan.html, Accessed on 30th March 2013.

[3] ChattyBot (2013), "ChattyBot – Free Chattbot for Android", Available at: http://sndapps.com/chattybot-free-chat-bot-for-android, Accessed on 08th April 2013.

[4] LPGenius (2013), "Chatterbot", Available at: http://www.calor.co.uk/customer-services, Accessed on 07th April 2013.

[5] Artificial Life3 (2012), "A Chat With Einstein", Available on: http://www.pandorabots.com/pandora/talk?botid=ea77c0200e365cfb, Accessed on 30th March 2013.

[6] R. Wallace (2003), *The Elements of AIML Style*, ALICE A.I Foundation, 2001.

[7] MakeAIML (2013) An AIML Creation Tool, Available at: http://makeaiml.aihub.org/tutorials/ra_aiml_basics.php, Accessed on 30th January 2013.

[8]     A. S. Lokman &, J. M. Zain (2010), "One-Match and All-Match for Keywords in Chatbot", *American Journal of Applied Sciences*, vol. 7, no 10, pp 1406-1411.

[9]     A. M. Turing, (1950), "Computing Machinery and Intelligence", *Mind*, vol. 59, pp. 433-460.

[10]    J. Weizenbaum (1966), "Eliza – a Computer Program for Study of Natural Communication Between Man and Machine", *Communications of the Association for Computing Machinery*, vol. 9, pp. 36-45.

[11]    M. L. Mauldin (1994), "Chatterbots, Tinymuds, and Turing Test: Entering the Loebner Prize Competition", *12th National Conference of the American Association for Artificial Intelligence*, pp. 16-21.

[12]    AliceBot (2013), AliceBot, Available at: http://alicebot.blogspot.com.br, Accessed on 30th January 2013.

[13]    E. A. Schegloff (2003), "Conversation Analysis and Communication Disorders", In C. Goodwin (Ed.), *Conversation and Brain Damage*, pp. 21-55, Oxford: Oxford University Press.

[14]    GaitoBot (2013), Aiml Chatbot Hosting, Available at: http://www.gaitobot.de/gaitobot/, Accessed on 20th January 2013.

[15]    Adeena Mignogna (2013), Simple Aiml Editor, Available at: http://riotsw.com/sae.html, Accessed on 20th January 2013.

[16]    N. Bush, (2013) Program D, Available at: http://aitools.org/Program_D, Accessed on 30th January 2013.

[17]    Program# (2013), Download Program# 2.0, Available at: http://aimlbot.sourceforge.net, Accessed on 20th January 2013.